\documentclass[letterpaper]{article} 
\pdfoutput=1 
\usepackage{aaai23}  
\usepackage{times}  
\usepackage{helvet}  
\usepackage{courier}  
\usepackage[hyphens]{url}  
\usepackage{graphicx} 
\usepackage{natbib}  
\usepackage{caption} 
\usepackage{algorithm}
\usepackage{algorithmic}
\usepackage{multirow}
\usepackage{float}
\usepackage{amssymb}
\nocopyright 
\usepackage{newfloat}
\usepackage{listings}
\DeclareCaptionStyle{ruled}{labelfont=normalfont,labelsep=colon,strut=off} 
\floatstyle{ruled}
\newfloat{listing}{tb}{lst}{}
\floatname{listing}{Listing}
\title{MNER-QG: An End-to-End MRC framework for Multimodal Named Entity Recognition with Query Grounding}
\author {
    Meihuizi Jia\textsuperscript{\rm 1,2},
    Lei Shen \textsuperscript{\rm 2},
    Xin Shen \textsuperscript{\rm 3},
    Lejian Liao \textsuperscript{\rm 1},
    Meng Chen \textsuperscript{\rm 2}\thanks{Corresponding author},
    Xiaodong He \textsuperscript{\rm 2},\\
    Zhendong Chen \textsuperscript{\rm 1},
    Jiaqi Li \textsuperscript{\rm 1}
}
\affiliations {
    \textsuperscript{\rm 1} School of Computer Science and Technology, Beijing Institute of Technology, Beijing, China\\
    \textsuperscript{\rm 2} JD AI Research, Beijing, China\\
    \textsuperscript{\rm 3} Australian National University,
Canberra, Australia\\
    \{jmhuizi24, liaolj, chenzd, 3120195492\}@bit.edu.cn, \{shenlei20, chenmeng20, xiaodong.he\}@jd.com, u6498962@anu.edu.au
}

\usepackage{bibentry}

\begin{document}

\maketitle

\begin{abstract}
Multimodal named entity recognition (MNER) is a critical step in information extraction, which aims to detect entity spans and classify them to corresponding entity types given a sentence-image pair. Existing methods either (1) obtain named entities with coarse-grained visual clues from attention mechanisms, or (2) first detect fine-grained visual regions with toolkits and then recognize named entities. However, they suffer from improper alignment between entity types and visual regions or error propagation in the two-stage manner, which finally imports irrelevant visual information into texts. In this paper, we propose a novel end-to-end framework named MNER-QG that can simultaneously perform MRC-based multimodal named entity recognition and query grounding. Specifically, with the assistance of queries, MNER-QG can provide prior knowledge of entity types and visual regions, and further enhance representations of both texts and images. To conduct the query grounding task, we provide manual annotations and weak supervisions that are obtained via training a highly flexible visual grounding model with transfer learning. We conduct extensive experiments on two public MNER datasets, Twitter2015 and Twitter2017. Experimental results show that MNER-QG outperforms the current state-of-the-art models on the MNER task, and also improves the query grounding performance.
\end{abstract}

\section{Introduction}

Multimodal named entity recognition (MNER) is a vision-language task 
that extends the traditional text-based NER and alleviates ambiguity in natural languages by taking images as additional inputs. 
The essence of MNER is to effectively capture visual features corresponding to entity spans and incorporate certain visual regions into textual representations.

\begin{figure}[ht]
  \includegraphics[scale=0.27]{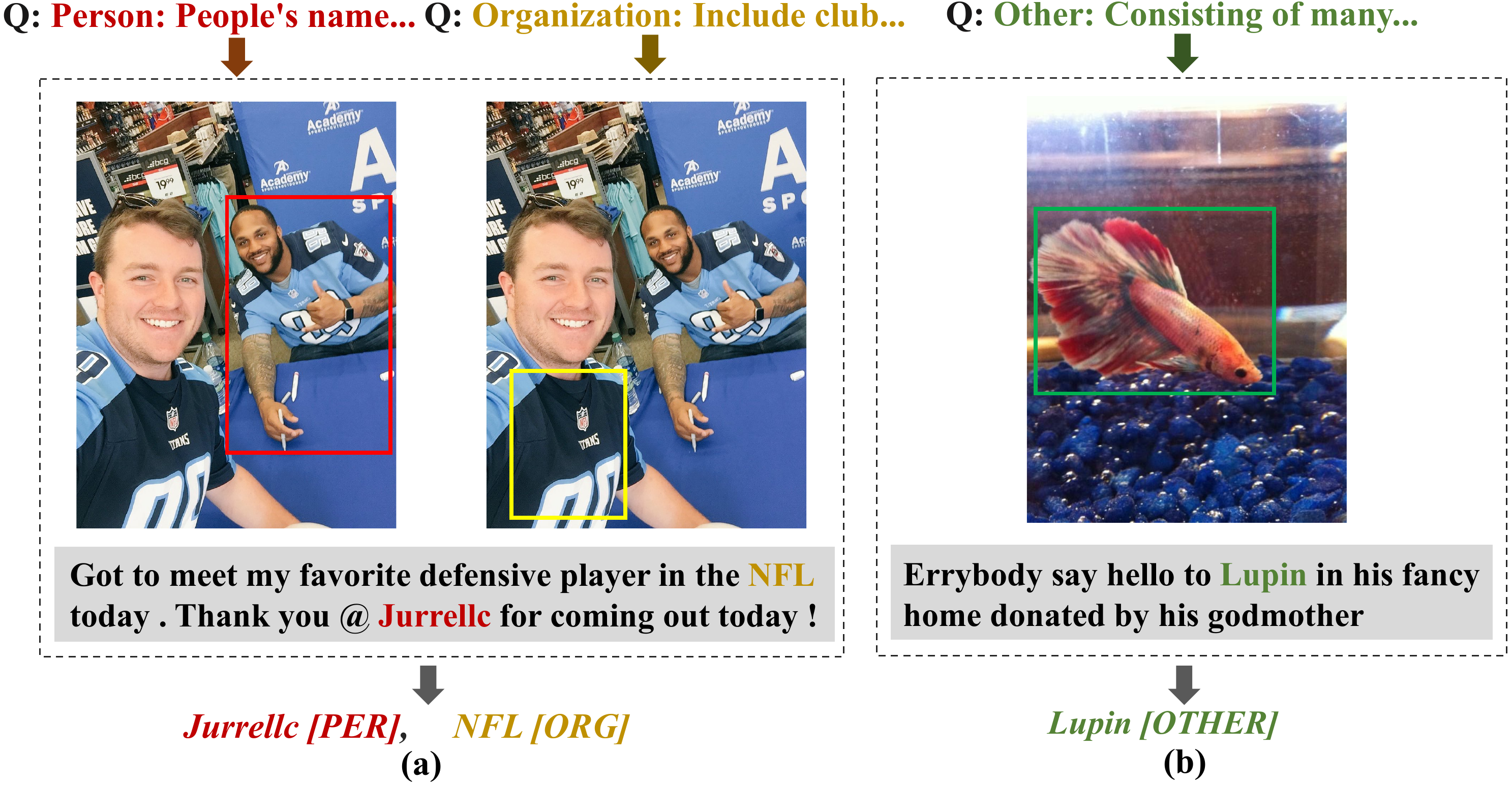}
  \caption{Two examples of MNER-QG with entity type ``ORG'', ''PER'', and ''OTHER''.}
  \label{fig:MRC-MNER process}
\end{figure}

Existing MNER datasets contain few fine-grained annotations in each sentence-image pair, i.e., the relevant image is given as a whole without regional signals for a particular entity type. 
Therefore, previous works implicitly align contents inside a sentence-image pair and fuse their representations based on various attention mechanisms \cite{moon2018multimodal,lu2018visual,zhang2018adaptive,arshad2019aiding,yu2020improving,chen2021multimodal,xu2022maf}. 
However, it is hard to interpret and evaluate the effectiveness of implicit alignments. 
Recently, visual grounding toolkits \cite{yang2019fast} are exploited to explicitly extract visual regions related to different entity types \cite{zhang2021multi}. 
The detected regions are then bound with the input sentence and fed into the recognition model together \cite{jia2022query}. 
Because of the two-stage manner, incorporating inaccurate visual regions from the first stage will hurt the final results (error propagation).

With respect to the problem formalization, early methods regard MNER as a sequence labeling task that integrates image embeddings into a sequence labeling model and assigns type labels to named entities. 
Recently, the machine reading comprehension (MRC) framework is employed in many natural language processing tasks due to its solid language understanding capability \cite{li2019unified,li2019entity,chen2021bidirectional}. 
To take advantage of the prior knowledge encoded in MRC queries \cite{li2019unified}, we consider MNER as a MRC task, which extracts entity spans by answering queries about entity types.
In addition, to capture the fine-grained alignment between entity types and visual regions, we ground the MRC queries on image regions and output their positions as bounding boxes.
For example, as shown in Figure 1 (a), recognizing entities with type PER and ORG in sentence ``Got to meet my favorite defensive player in the NFL today. Thank you @ Jurrellc for coming out today!'' is formalized as extracting answer spans from the input sentence given the query ``Person: People's name...'' and ``Organization: Include club...''. 
Then, answer spans ``Jurrellc'' and ``NFL'' are obtained along with their visual regions marked by red and yellow boxes.

To this end, we propose an end-to-end MRC framework for \textbf{M}ultimodal \textbf{N}amed \textbf{E}ntity \textbf{R}ecognition with \textbf{Q}uery \textbf{G}rounding (\textbf{MNER-QG}).
This joint-training approach forces the model to explicitly align entity spans with the corresponding visual regions, and further improves the performance of both named entity recognition and query grounding.
Specifically, we design unified queries with prior information as navigators to pilot our joint-training model. 
Meanwhile, we extract multi-scale visual features and design two interaction mechanisms, multi-scale cross-modality interaction and existence-aware uni-modality interaction, to enrich both textual and visual information. 
Since there are few fine-grained annotations for visual regions in existing MNER datasets, we provide two types of bounding box annotations, weak supervisions and manual annotations. 
The former is obtained by training a visual grounding model with transfer learning, while the latter aims to provide oracle results.


In summary, the contribution of this paper is three-fold:
\begin{itemize}
    \item We propose a novel end-to-end MRC framework, MNER-QG. Our model simultaneously performs MRC-based multimodal named entity recognition and query grounding in a joint-training manner. To the best of our knowledge, this is the first attempt on MNER.
    
    \item To fulfill the end-to-end training, we contribute weak supervisions via training a visual grounding model with transfer learning. Meanwhile, we offer manual annotations of bounding boxes as oracle results.
    
    
    \item We conduct extensive experiments on two public MNER datasets, Twitter2015 and Twitter2017, to evaluate the performance of our framework. Experimental results show that MNER-QG outperforms the current state-of-the-art models on both datasets for MNER, and also improves the QG performance.
    
   
\end{itemize}

\section{Related Work}

\subsection{Multimodal Named Entity Recognition}

With the increasing popularity of multimodal data on social media platforms, multimodal named entity recognition (MNER) has become an important research direction, which assists the NER models \cite{modularized,effective,unified} in better identifying entities by taking images as the auxiliary input. 
The critical challenge of MNER is how to align and fuse textual and visual information. 
\citet{yu2020improving} proposed a multimodal transformer architecture for MNER, which captures expressive text-image representations by incorporating the auxiliary entity span detection.
\citet{zhang2021multi} created the graph connection between textual words and visual objects acquired by a visual grounding toolkit \cite{yang2019fast}, and proposed a graph fusion approach to conduct graph encoding. 
\citet{xu2022maf} proposed a matching and alignment framework for MNER to improve the consistency of representations in different modalities. 

Lacking prior information of entity types and accurate annotations of visual regions corresponding to certain entity types, the above methods feed visual information (an entire image, image patches, or retrieved visual regions from toolkits) with the entire sentence into an entity recognition model, which inevitably makes it difficult to obtain the explicit alignment between images and texts.

\subsection{Machine Reading Comprehension}
Machine Reading Comprehension (MRC) aims to answer natural language queries given a set of contexts where the answers to these queries can be inferred. 
In various forms of MRC, span extraction MRC \cite{peng2021aper,jia2022keywords} is challenging, which extracts a span as the answer from context. 
The span extraction can be regarded as two multi-class classification or two binary classification tasks. 
For the former, the model needs to predict the start and end positions of an answer. 
For the latter, the model needs to decide whether each token is the start/end position. 
Recurrent Neural Network (RNN) was used to encode textual information, then a linear projection layer was adopted to predict answer spans  
\cite{yang2018hotpotqa, nishida2019answering}.
The performance was boosted with the development of large-scale pre-trained models \cite{qiu2019dynamically, tu2020select}, such as ELMo \cite{PetersNIGCLZ18}, BERT \cite{kenton2019bert}, and RoBERTa \cite{liu2019roberta}.

Recently, there is a trend of converting NLP tasks to the MRC form, including named entity recognition \cite{li2019unified}, entity relation extraction \cite{li2019entity}, and sentiment analysis \cite{chen2021bidirectional}. Due to the powerful understanding ability contained in MRC, the model performance of these tasks is improved. 


\subsection{Visual Grounding}
Visual grounding aims to localize textual entities or referring expressions in an image.
This task is divided into two paradigms: two-stage and one-stage. For the former, the first stage is exploited to extract region proposals as candidates via some region proposal methods (e.g., Edgebox \cite{zitnick2014edge}, selective search \cite{uijlings2013selective}, and Region Proposal Networks \cite{ren2015faster}), and then the second stage is designed to rank region-text candidate pairs. 
\begin{figure*}[ht]
  \centering
  \includegraphics[scale=0.48]{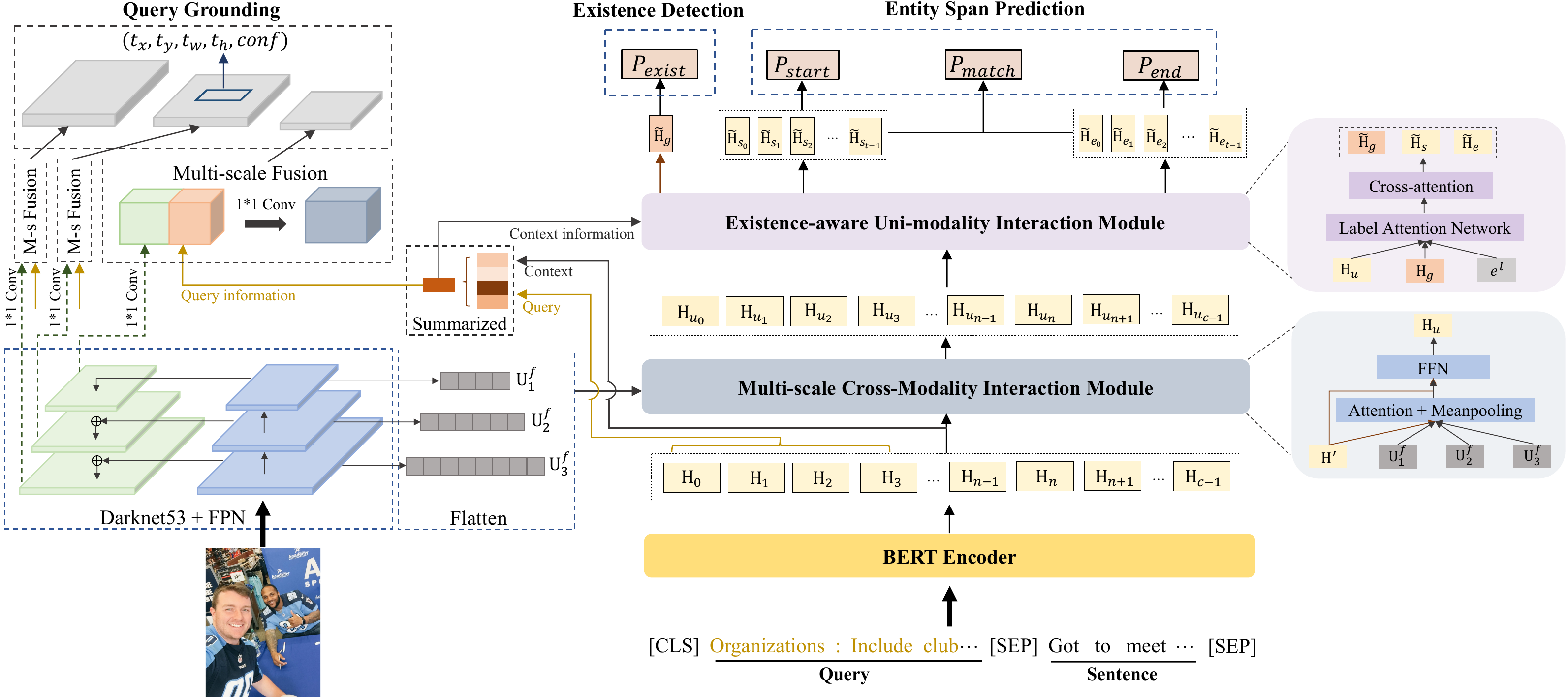}
  \caption{Overview of our MNER-QG framework (M-s Fusion denotes Multi-scale Fusion).}
  \label{fig:MNER-QG process}
\end{figure*}
For the latter, researchers utilize one-stage model (e.g., YOLO \cite{redmon2018yolov3,bochkovskiy2020yolov4}) combined with extra features to directly output the final region(s). 
Compared with the two-stage manner, the one-stage framework is simplified and accelerates the inference by conducting detection and matching simultaneously.

To connect visual grounding and MRC-based named entity recognition effectively, we use queries from MRC as input texts and force model to perform query grounding. Since queries contain the prior knowledge of entity types, our work can achieve the explicit alignment between entity types and visual regions.


\section{METHOD}
\label{sec:method}

\subsection{Overview}

Figure \ref{fig:MNER-QG process} illustrates the overall architecture of MNER-QG. 
Given a sentence $S = \left\{ s_{0},s_{1},...,s_{n-1}\right\}$ and its associated image $V$, where $n$ denotes the sentence length, we first design a natural language query $Q = \left\{ q_{0}, q_{1},...,q_{m-1}\right\}$ with prior awareness about entity types. 
Then, our model performs multi-scale cross-modality interaction and existence-aware uni-modal interaction to simultaneously detect entity spans $s_{\mathrm{start,end}}$ and the corresponding visual regions via answering the query $Q$. 




\subsection{Query Construction}
Query plays a significant role as the navigator in our MNER-QG, and it should be expressed as generic, precise, and effective as possible. 
Table \ref{tab:query construction} shows examples of queries designed by us. 
We hope that the queries are moderate in difficulty and can provide informative knowledge of MNER and QG tasks. 
Therefore, the model can stimulate the solid understanding capability of MRC without limiting the performance of QG.

\begin{table}[ht]
\small
\begin{tabular}{lp{4.9cm}}
\hline
\textbf{Entity Type} & \textbf{Natural Language Query} \\ \hline
PER (Person) & Person: People's name and fictional character. \\ \hline
\multirow{2}*{LOC (Location)} & Location: Country, city, town continent by geographical location. \\ \hline
\multirow{3}*{ORG (Organization)} & Organization: Include club, company, government party, school government, and news organization. \\ \hline
\end{tabular}
\caption{Examples of transforming entity types to queries.}
\label{tab:query construction}
\vspace{-2mm}
\end{table}

\subsection{Model Architecture}
\label{sub:Text-Visual Encoding and Unifying}

\subsubsection{Input Representation.}
For text information, we concatenate a query and sentence pair $\{{\mathrm{[CLS]},Q,\mathrm{[SEP]},S,\mathrm{[SEP]}}\}$, 
and encode the result to a 768$d$ real-valued vector with the pre-trained BERT model \cite{kenton2019bert}, where $\mathrm{[CLS]}$ and $\mathrm{[SEP]}$ are special tokens. 
Then BERT outputs a contextual representation $\mathbf{H} \in \mathbb{R}^{c\times d_{c}}$, where $c=m+n+3$ is the length of BERT input. 
For visual information, inspired by \citet{yang2019fast}, we use Darknet-53 \cite{zhu2016visual7w} with feature pyramid networks \cite{lin2017feature} to extract visual features. 
The images are resized to 256 $\times$ 256, and the feature maps are $\frac{1}{32}$, $\frac{1}{16}$, and $\frac{1}{8}$, respectively. 
Therefore, the three spatial resolutions are $8 \times 8 \times d_{1}$, $16 \times 16 \times d_{2}$, and $32 \times 32 \times d_{3}$, where $d_{1}=1024$, $d_{2}=512$, and $d_{3}=256$ are feature channels. 


We unify the dimensions of three visual features and textual feature to facilitate the model computation. 
Specifically, we add a $1 \times 1$ convolution layer with batch normalization and RELU under the feature pyramid networks to map the feature channels $d_{1}$, $d_{2}$, and $d_{3}$ to the same dimension $d=512$.
The new visual features are denoted as $\mathbf{U}_{1}$, $\mathbf{U}_{2}$, and $\mathbf{U}_{3}$.
At the same time, we flatten $8 \times 8$, $16 \times 16$, and $32 \times 32$ to 64, 256, and 1024, which are used to generate the visual representations, $\mathbf{U}_1^f$, $\mathbf{U}_2^{f}$ and $\mathbf{U}_3^f$.
For textual information, we use a linear projection to map $d_{c}$ to $d=512$, and the mapped representation is $\mathbf{H}^{'}$.

\subsubsection{Multi-scale Cross-modality Interaction.}
This module is shown in Figure \ref{fig:MNER-QG process} with the grey box. 
We first truncate the token-level representations of query $\mathbf{Q}$ from $\mathbf{H}^{'}$.
The query encoding now contains messages from the original sentence $S$, which can be passed to the QG task.
Then, we use an attention-based approach \cite{rei2019jointly} to acquire the summarized query representation $\mathbf{q}\in \mathbb{R}^{1\times d}$ that will be fed into QG.
\begin{equation}
\label{attention-based summarized approach}
\alpha=\mathrm{softmax} \left ( \mathrm{MLP}\left(\mathbf{Q}\right)\right),\quad \mathbf{q}=\sum_{k=0}^{m-1}\alpha_{k}\mathbf{Q}\left [ k,: \right ]  
\end{equation}

To fully exploit image information, we employ multi-scale visual representations to update textual representation through a cross-modality attention mechanism, where $\mathbf{H}^{'}$ works as the query matrix, while each of $\mathbf{U}_{1}^{f}$, $\mathbf{U}_{2}^{f}$, and $\mathbf{U}_{3}^{f}$ works as the key and value matrix.
The visual-enhanced attention outputs are denoted as $\mathbf{H}_{1}$, $\mathbf{H}_{2}$, and $\mathbf{H}_{3} \in \mathbb{R}^{c\times d}$. 
Then, we merge these matrices to a unified textual representation $\mathbf{H}_{a}$ using $\mathrm{MeanPooling}$.
Finally, we concatenate $\mathbf{H}_{a}$ and $\mathbf{H}^{'}$, and feed the result into a feed-forward neural network to get the final textual representation $\mathbf{H}_{u}$.

\subsubsection{Existence-aware Uni-modality Interaction.}
\label{sub:Existence-aware Uni-modal Interaction}
Since the sentence does not always contain the entity asked by the current query, we design a global existence signal to enhance the model's awareness of entity existence.
Similar to Equation (\ref{attention-based summarized approach}), we summarize contextual representation $\mathbf{H}^{'}$ to acquire the existence representation $\mathbf{H}_{g}\in\mathbb{R}^{1\times d}$. 
Inspired by \citet{qin2021co} and \citet{modularized}, we then employ a label attention network to update both the textual representation with the encoding of start/end label and the existence representation with the encoding of existence label. (Note that the $\mathbf{e}^{l}$ in Figure \ref{fig:MNER-QG process} denotes a label embedding lookup table). Details of the label attention network are provided in the Appendix.
Then, we get start/end label-enhanced textual representation, $\mathbf{H}_{s}$/$\mathbf{H}_{e}$, which can be regarded as the start/end representation of entity span, and also label-enhanced existence representation $\mathbf{\widehat{H}}_{g}$.

We calculate attention scores between $\mathbf{H}_{s}$ and $\mathbf{\widehat{H}}_{g}$, and define the existence-aware start representation $\mathbf{\widetilde{H}}_{s}$ as follows:
\begin{equation}
\mathbf{Z}_{s} = \mathrm{softmax}\left ( \frac{\mathbf{Q}_{s}\mathbf{K}_{g}^{\top }}{\sqrt{d_{k}} }  \right ) \mathbf{V}_{g},
\mathbf{\widetilde{H}}_{s} = \mathrm{LN}\left ( \mathbf{H}_{s} + \mathbf{Z}_{s} \right ),
\end{equation}
where LN denotes the layer normalization function \cite{ba2016layer}, $\mathbf{Q}_{s} \in \mathbb{R}^{c\times d}$, $\mathbf{K}_{g}, \mathbf{V}_{g} \in \mathbb{R}^{1\times d}$, and $\mathbf{\widetilde{H}}_{s} \in \mathbb{R}^{c\times d}$. 
Similarly, we can obtain the updated end representation $\mathbf{\widetilde{H}}_{e} \in \mathbb{R}^{c\times d}$, and the updated existence representation $\mathbf{\widetilde{H}}_{g} \in \mathbb{R}^{1\times d}$.

\subsection{Query Grounding}

Following \citet{yang2019fast}, we first broadcast the query representation $\mathbf{q}$ 
to each spatial location, denoted as $\left ( i,j \right ) $, and then concatenate the query feature and visual feature $\mathbf{U}_{i}$, where $i=1,2,3$. 
The feature dimension after concatenation is $512+512=1024$. 
Another $1 \times 1$ convolution layer is appended to better fuse above features at each location independently and map them to the dimension $d=512$. 

Next, we perform the grounding operation. 
There are $8\times 8+16\times 16+32\times 32=1344$ locations in three spatial resolutions, and each location is related to a $512D$ feature vector from the fusion layer. 
YOLOv3 network centers around each of the location's three anchor boxes, hence it predicts bounding boxes at three scales. 
The output of YOLOv3 network is $\left [ 3\times\left ( 4+1 \right )  \right ] \times r_{i}\times r_{i}$ at each scale for shifting the center, width, and height $\left ( t_{x},t_{y},t_{w},t_{h}\right ) $ of the anchor box, along with the confidence score on this shifted box, where $r_{i}\times r_{i}$ denotes the shape size of each spatial resolution. 
Ultimately, only one region is desired as the output for query grounding. 
More details can be found in \citet{yang2019fast}.

The objective function $\mathcal{L}_{QG}$ of QG task consists of regression loss on bounding box $\mathcal{L}_{bbox}$ and objectness score loss $\mathcal{L}_{object}$.
$\mathcal{L}_{bbox}$ is expected to assign bounding box regions to ground truth objects via mean squared error (MSE). $\mathcal{L}_{object}$ is used to classify the bounding box regions as object or non-object via binary cross-entropy (BCE).

\subsection{Multimodal Named Entity Recognition}
The core of multimodal named entity recognition is to predict the entity span in sentence. In this section, we design an auxiliary task named existence detection (ED) after receiving the existence representation $\mathbf{\widetilde{H}}_{g}$ to predict whether a sentence contains entities with specific type and cooperate with the entity span prediction task to extract entity span.
\subsubsection{Existence Detection.} This task and the entity span prediction task can share the corresponding mutual information with the co-interactive attention mechanism. The existence of entity is detected as follows:
\begin{equation}
\mathrm{P}_{exist} = \mathrm{softmax} \left ( \mathbf{\widetilde{H}}_{g} \mathbf{W}_{exist}  \right ) 
\end{equation}
where $\mathbf{W}_{exist} \in \mathbb{R}^{d\times 2}$ and $\mathrm{P}_{exist}\in \mathbb{R}^{1\times 2}$. We formulate the ED sub-task as a text classification task. The loss function is denoted by $\mathcal{L}_{ED}$ and the binary cross-entropy (BCE) loss is taken as the training objective.

\subsubsection{Entity Span Prediction.}
To tag the entity span from a sentence using MRC framework, it is necessary to find the start and end positions of the entity. We utilize two binary classifiers to predict whether each token in the sentence is the start/end index or not, respectively. The probability that each token is predicted to be a start position is as follows:
\begin{equation}
\mathrm{P}_{start} = \mathrm{softmax_{each row}}\left ( \mathbf{\widetilde{H}}_{s} \mathbf{W}_{s} \right )
\end{equation}
where $\mathbf{W}_{s} \in \mathbb{R}^{d \times 2}$ and $\mathrm{P}_{start} \in \mathbb{R}^{c \times 2}$. Similarly, we can get the probability of the end position $\mathrm{P}_{end} \in  \mathbb{R}^{c \times 2}$.

Since there could be multiple entities of the same type in the sentence, we add a binary classification model to predict the matching probability of start and end positions inspired by \citet{li2019unified}.
\begin{equation}
\mathrm{P}_{match} = \mathrm{sigmoid}\left ( \mathbf{W}_{m}\left [ \mathbf{\widetilde{H}}_{s};\mathbf{\widetilde{H}}_{e} \right ] \right )   
\end{equation}
where $\mathbf{W}_{m}\in \mathbb{R}^{1 \times 2d}$, $\mathrm{P}_{match} \in  \mathbb{R}^{1 \times 2}$. $\left[ ;\right]$ is denoted the concatenation in columns. 

During training, the objective function $\mathcal{L}_{ESP}$ of entity span prediction (ESP) sub-task consists of start position loss $\mathcal{L}_{start}$, end position loss $\mathcal{L}_{end}$ and matching loss $\mathcal{L}_{match}$, where binary cross-entropy (BCE) is used for calculation.

Finally, combining the two tasks QG and MNER, the overall objective function is as follows:
\begin{equation}
\mathcal{L}=\omega_{f} \mathcal{L}_{QG}+\lambda_{1}\mathcal{L}_{ED}+\lambda_{2}\mathcal{L}_{ESP}
\end{equation}
where $\omega_{f},\mathcal{\lambda}_{2}$ and $\mathcal{\lambda}_{3}$ are hyper-parameters to control the contributions of each sub-task.

\section{Experiments}

\subsection{Dataset Construction} 
There are two widely-used MNER datasets, Twitter2015 \cite{zhang2018adaptive} and Twitter2017 \cite{lu2018visual}, used to evaluate the effectiveness of our framework. Both datasets are separated into training, validation, and test sets with the same type distribution. 
Statistics are listed in Appendix. And then, we contribute two types of labels: weak supervisions and manual annotations for public research. 

For weak supervisions, we apply the pre-trained fast and accurate one-stage visual grounding model \cite{yang2019fast} (denoted as FA-VG) as the base model. In the setting of Phrase Localization task, FA-VG was trained and evaluated on the Flickr30K Entities dataset \cite{plummer2015flickr30k} that augments the original Flickr30K \cite{young2014image} with region-phrase correspondence annotations. 
However, there are two obstacles: (1) These phrases/queries are from image captions, and not specially constructed for the named entity recognition task. (2) The MNER datasets (i.e. Twitter2015/2017) have different data domains compared with the Flickr30K Entities dataset. Thus, we utilize transfer learning to overcome above issues. 
In addition, we contribute manual annotations for public research. We hire three crowd-sourced workers who are familiar with the tasks of MNER and target detection to help us annotate the bounding box in the image. The annotators are requested to tag the visual regions in the image corresponding to the entity span in the sentence. After the data annotation, we merge the instances of strong inter-annotator agreement from three crowd-sourced workers to acquire high-quality and explicit text-image alignment data. Details of the annotation with two types of labels are provided in the Appendix.



\subsection{Experiment Settings}
\subsubsection{Evaluation Metrics.} 
For the MNER task, we use precision ($Pre.$), recall ($Rec.$), and F1 score ($F1$) to evaluate the performance of overall entity types, and use $F1$ only for each type. For the QG, we follow prior works \cite{rohrbach2016grounding} and utilize Accu@0.5 as evaluation protocol. Given a query, an output image region is considered correct if its IoU is at least 0.5 with the ground truth bounding box. In addition, we add Accu@0.75 (IoU is at least 0.75) and Miou (mean of IoU) as additional evaluation metrics.

\subsubsection{Implementation Details.} 
The learning rate and dropout rate are set to 5e-5 and 0.3, which obtains the best performance on the validation set of two datasets after conducting a grid search over the interval [1e-5, 1e-4] and [0.1, 0.6]. 
We train the model with AdamW optimization. 
To further evaluate our joint-training model, we take out the images from Twitter2015/2017 to train the QG model separately. For a fair comparison, we use the same configurations such as batch size, learning rate, and optimizer in both the QG model and our joint-training model. For the joint-training loss, we set the hyper-parameters $\mathcal{\lambda}_{1}=1$ and $\mathcal{\lambda}_{2}=2$ by tuning on the validation set. 
We specially set a balance factor $\omega_{f}$ to dynamically scale the loss of MNER and QG. Please refer to Appendix for calculation details. 

\subsubsection{Baseline Models.}
Two groups of baselines are compared with our approach. The first group consists of some text-based MNER models that formalize MNER as a sequence labeling task:
(1) \textbf{BiLSTM-CRF} \cite{huang2015bidirectional};
(2) \textbf{CNN-BiLSTM-CRF} \cite{ma2016end}; 
(3) \textbf{HBiLSTM-CRF} \cite{lample2016neural}; 
(4) \textbf{BERT} \cite{kenton2019bert};
(5) \textbf{BERT-CRF}; 
(6) \textbf{T-NER} \cite{ritter2011named,zhang2018adaptive}. 
The second group contains several competitive MNER models: 
(1) \textbf{GVATT-HBiLSTM-CRF} \cite{lu2018visual};
(2) \textbf{GVATT-BERT-CRF} \cite{yu2020improving};
(3) \textbf{AdaCAN-CNN-BiLSTM-CRF} \cite{zhang2018adaptive};
(4) \textbf{AdaCAN-BERT-CRF} \cite{yu2020improving};
(5) \textbf{UMT-BERT-CRF} \cite{yu2020improving};
(6) \textbf{MT-BERT-CRF} \cite{yu2020improving}; 
(7) \textbf{ATTR-MMKG-MNER} \cite{chen2021multimodal}; 
(8) \textbf{UMGF} \cite{zhang2021multi}; 
(9) \textbf{MAF} \cite{xu2022maf}. 
The details of these models are illustrated in Appendix.

According to different derivations of bounding box labels in the images, we provide two versions of our model \textbf{MNER-QG} and \textbf{MNER-QG (Oracle)} for evaluation. 
In addition, we provide a variant of the model, \textbf{MNER-QG-Text}, which uses text input only.

\subsection{Main Results}

Table \ref{tab:Performance Comparison on Two MNER Datasets} shows the results of our model and baselines. The upper results are from text-based models and the lower results are from multimodal models. Firstly, we compare the multimodal models with their corresponding uni-modal baselines in MNER, such as AdaCAN-CNN-BiLSTM-CRF vs. CNN-BiLSTM-CRF, and MNER-QG vs. MNER-QG-Text. We notice almost all multimodal models can significantly outperform their corresponding uni-modal competitors, 
indicating the effectiveness of images. And then, we compare our MNER-QG with other multimodal baselines. The result shows MNER-QG outperforms all baselines on Twitter2017 and yields competitive results on Twitter2015. MNER-QG (Oracle) with more accurate manual annotations yields further results in both datasets.


\begingroup
\setlength{\tabcolsep}{3.5pt}
\renewcommand{\arraystretch}{0.9}
\begin{table*}[ht]
\small
\centering
\begin{tabular}{c|cccccccccccccc}
\hline
  \multirow{3}{*}{Methods} &
  \multicolumn{7}{c|}{Twitter2015} &
  \multicolumn{7}{c}{Twitter2017} \\ \cline{2-15}
 &
  \multicolumn{4}{c|}{Single Type (\textit{F1})} &
  \multicolumn{3}{c|}{Overall} &
  \multicolumn{4}{c|}{Single Type (\textit{F1})} &
  \multicolumn{3}{c}{Overall} \\
   &
  \textbf{PER} &
  \textbf{LOC} &
  \textbf{ORG} &
  \multicolumn{1}{c|}{\textbf{OTH.}} &
  \textit{Pre}. &
  \textit{Rec.} &
  \multicolumn{1}{c|}{\textit{F1}} &
  \textbf{PER} &
  \textbf{LOC} &
  \textbf{ORG} &
  \multicolumn{1}{c|}{\textbf{OTH.}} &
  \textit{Pre.} &
  \textit{Rec.} &
  \textit{F1} \\ \hline
BiLSTM-CRF &
  76.77 &
  72.56 &
  41.33 &
  \multicolumn{1}{c|}{26.80} &
  68.14 &
  61.09 &
  \multicolumn{1}{c|}{64.42} &
  85.12 &
  72.68 &
  72.50 &
  \multicolumn{1}{c|}{52.56} &
  79.42 &
  73.43 &
  76.31 \\
CNN-BiLSTM-CRF &
  80.86 &
  75.39 &
  47.77 &
  \multicolumn{1}{c|}{32.61} &
  66.24 &
  68.09 &
  \multicolumn{1}{c|}{67.15} &
  87.99 &
  77.44 &
  74.02 &
  \multicolumn{1}{c|}{60.82} &
  80.00 &
  78.76 &
  79.37 \\
HBiLSTM-CRF &
  82.34 &
  76.83 &
  51.59 &
  \multicolumn{1}{c|}{32.52} &
  70.32 &
  68.05 &
  \multicolumn{1}{c|}{69.17} &
  87.91 &
  78.57 &
  76.67 &
  \multicolumn{1}{c|}{59.32} &
  82.69 &
  78.16 &
  80.37 \\
BERT &
  84.72 &
  79.91 &
  58.26 &
  \multicolumn{1}{c|}{38.81} &
  68.30 &
  \textbf{74.61} &
  \multicolumn{1}{c|}{71.32} &
  90.88 &
  84.00 &
  79.25 &
  \multicolumn{1}{c|}{61.63} &
  82.19 &
  83.72 &
  82.95 \\
BERT-CRF &
  \textbf{84.74} &
  80.51 &
  \textbf{60.27} &
  \multicolumn{1}{c|}{37.29} &
  69.22 &
  74.59 &
  \multicolumn{1}{c|}{71.81} &
  90.25 &
  83.05 &
  81.13 &
  \multicolumn{1}{c|}{62.21} &
  83.32 &
  83.57 &
  83.44 \\
T-NER &
  83.64 &
  76.18 &
  59.26 &
  \multicolumn{1}{c|}{34.56} &
  69.54 &
  68.65 &
  \multicolumn{1}{c|}{69.09} &
  - &
  - &
  - &
  \multicolumn{1}{c|}{-} &
  - &
  - &
  - \\ 
  \textbf{MNER-QG-Text (Ours)} &
  84.72 &
  \textbf{81.13} &
  60.07 &
  \multicolumn{1}{c|}{\textbf{39.23}} &
  \textbf{76.35} &
  {69.46} &
  \multicolumn{1}{c|}{\textbf{72.74}} &
  \textbf{91.33} &
  \textbf{85.23} &
  \textbf{81.75} &
  \multicolumn{1}{c|}{\textbf{68.41}} &
  \textbf{87.12} &
  \textbf{84.03} &
  \textbf{85.55} \\\hline
GVATT-HBiLSTM-CRF &
  82.66 &
  77.21 &
  55.06 &
  \multicolumn{1}{c|}{35.25} &
  73.96 &
  67.90 &
  \multicolumn{1}{c|}{70.80} &
  89.34 &
  78.53 &
  79.12 &
  \multicolumn{1}{c|}{62.21} &
  83.41 &
  80.38 &
  81.87 \\
AdaCAN-CNN-BiLSTM-CRF &
  81.98 &
  78.95 &
  53.07 &
  \multicolumn{1}{c|}{34.02} &
  72.75 &
  68.74 &
  \multicolumn{1}{c|}{70.69} &
  89.63 &
  77.46 &
  79.24 &
  \multicolumn{1}{c|}{62.77} &
  84.16 &
  80.24 &
  82.15 \\
GVATT-BERT-CRF &
  84.43 &
  80.87 &
  59.02 &
  \multicolumn{1}{c|}{38.14} &
  69.15 &
  74.46 &
  \multicolumn{1}{c|}{71.70} &
  90.94 &
  83.52 &
  81.91 &
  \multicolumn{1}{c|}{62.75} &
  83.64 &
  84.38 &
  84.01 \\
AdaCAN-BERT-CRF &
  85.28 &
  80.64 &
  59.39 &
  \multicolumn{1}{c|}{38.88} &
  69.87 &
  74.59 &
  \multicolumn{1}{c|}{72.15} &
  90.20 &
  82.97 &
  82.67 &
  \multicolumn{1}{c|}{64.83} &
  85.13 &
  83.20 &
  84.10 \\
MT-BERT-CRF &
  85.30 &
  81.21 &
  61.10 &
  \multicolumn{1}{c|}{37.97} &
  70.84 &
  74.80 &
  \multicolumn{1}{c|}{72.58} &
  91.47 &
  82.05 &
  81.84 &
  \multicolumn{1}{c|}{65.80} &
  84.60 &
  84.16 &
  84.42 \\
UMT-BERT-CRF &
  85.24 &
  81.58 &
  63.03 &
  \multicolumn{1}{c|}{39.45} &
  71.67 &
  \textbf{75.23} &
  \multicolumn{1}{c|}{73.41} &
  91.56 &
  84.73 &
  82.24 &
  \multicolumn{1}{c|}{70.10} &
  85.28 &
  85.34 &
  85.31 \\
ATTR-MMKG-MNER &
  84.28 &
  79.43 &
  58.97 &
  \multicolumn{1}{c|}{41.47} &
  74.78 &
  71.82 &
  \multicolumn{1}{c|}{73.27} &
  - &
  - &
  - &
  \multicolumn{1}{c|}{-} &
  - &
  - &
  - \\
UMGF &
  84.26 &
  \textbf{83.17} &
  62.45 &
  \multicolumn{1}{c|}{\textbf{42.42}} &
  74.49 &
  75.21 &
  \multicolumn{1}{c|}{\textbf{74.85}} &
  91.92 &
  85.22 &
  83.13 &
  \multicolumn{1}{c|}{69.83} &
  86.54 &
  84.50 &
  85.51 \\
MAF &
  84.67 &
  81.18 &
  63.35 &
  \multicolumn{1}{c|}{41.82} &
  71.86 &
  75.10 &
  \multicolumn{1}{c|}{73.42} &
  91.51 &
  85.80 &
  \textbf{85.10} &
  \multicolumn{1}{c|}{68.79} &
  86.13 &
  \textbf{86.38} &
  86.25 \\
\textbf{\textbf{MNER-QG (Ours)}} &
  \textbf{85.31} &
  81.65 &
  \textbf{63.41} &
  \multicolumn{1}{c|}{41.32} &
  \textbf{77.43} &
  72.15 &
  \multicolumn{1}{c|}{74.70} &
  \textbf{92.92} &
  \textbf{86.19} &
  84.52 &
  \multicolumn{1}{c|}{\textbf{71.67}} &
  \textbf{88.26} &
  85.65 &
  \textbf{86.94} \\ 
\textbf{\textbf{MNER-QG (Oracle) (Ours)}} &
  \textbf{85.68} &
  81.42 &
  \textbf{63.62} &
  \multicolumn{1}{c|}{41.53} &
   \textbf{77.76} &
   72.31 &
  \multicolumn{1}{c|}{\textbf{74.94}} &
  \textbf{93.17} &
  \textbf{86.02} &
  84.64 &
  \multicolumn{1}{c|}{\textbf{71.83}} &
  \textbf{88.57} &
  85.96 &
  \textbf{87.25} \\ \hline
\end{tabular}
\caption{Results on two MNER datasets. We refer to the results of UMGF from \citet{zhang2021multi} and other results from \citet{xu2022maf}. Our model achieves a statistically significant improvement with p-value$<$0.05 under a paired two-sided t-test.}
\label{tab:Performance Comparison on Two MNER Datasets}
\end{table*}

\begingroup
\setlength{\tabcolsep}{3pt}
\renewcommand{\arraystretch}{0.9}
\begin{table}[ht]
\small
\centering
\begin{tabular}{l|ccc|ccc}
\hline
\multicolumn{1}{c|}{\multirow{2}{*}{Methods}} & \multicolumn{3}{c|}{Twitter2015} & \multicolumn{3}{c}{Twitter2017} \\
\multicolumn{1}{c|}{} & \textit{Pre.} & \textit{Rec.} & \textit{F1} & \textit{Pre.} & \textit{Rec.} & \textit{F1} \\ \hline
\multicolumn{1}{c|}{MNER-QG} & 77.43 & 72.15 & 74.70 & 88.26 & 85.65 & 86.94 \\ \hline
\quad - w/o QG loss & 77.50 & 70.79 & 73.99 & 88.01 & 84.69 & 86.32 \\
\quad - w/o ED loss & 77.53 & 71.20 & 74.23 & 87.81 & 85.28 & 86.53  \\
\quad - w/o QG+ED loss & 77.17 & 70.29 & 73.57 & 87.63 & 84.47 & 86.02 \\ \hline
\end{tabular}
\caption{Ablation study of MNER-QG on Test set.}
\label{tab:ablation}
\end{table}

\subsection{Ablation Study}


Table \ref{tab:ablation} shows the ablation results. We observe that all sub-tasks are necessary.
First, after removing the QG loss, the performance significantly drops on all metrics. In particular, $F1$ scores on two datasets degrade by 0.71\% and 0.62\%, respectively. The result shows the QG training promotes explicit alignment between text and image. Besides, removing the ED loss also damages the performance on all metrics. $F1$ scores on the two datasets decrease by 0.47\% and 0.41\%, respectively. We conjecture that ED provides global information for the entire model, which can help the model determine whether the sentence contains certain entities asked by the query. Finally, after removing both QG and ED loss, the performance drops more significantly, indicating that both the QG and ED tasks are essential in our framework.




\subsection{Case Study}
Here we conduct further qualitative analysis with two specific examples.
We compare the results from MNER-QG, MNER-QG (Oracle), and the competitive model UMGF. In Figure \ref{fig:case_study} (a), the sentence contains two entities ``\textit{lebron james}'', and ``\textit{Cavaliers}'' with PER, and ORG types respectively. However, UMGF locates the entity ``\textit{lebron james}'' inaccurately and misjudge its type. We guess it is because UMGF cannot detect the region of person on the red T-shirt. Instead, both MNER-QG and MNER-QG (Oracle) extract region about ``\textit{lebron james}'' (red box) for PER, and the logo about ``\textit{Cavaliers}'' (yellow box) for ORG on clothing, and the regions extracted by MNER-QG (Oracle) are more accurate due to the more elaborate manual annotations. Compared with UMGF, our model can locate more relevant visual regions, which can assist the model on accurately recognizing entities. Figure \ref{fig:case_study} (b) shows a more challenging case, where the image cannot provide useful regions about LOC. It can be seen that UMGF, MNER-QG and MNER-QG (Oracle) cannot locate the relevant visual regions for this entity. However, both MNER-QG and MNER-QG (Oracle) can recognize ``\textit{Epcot}'' and its type. We conjecture that the solid understanding capability of MRC and the guidance of query prior information contribute to the final correct prediction.
\begin{figure*}[ht]
  \centering
  \includegraphics[scale=0.46]{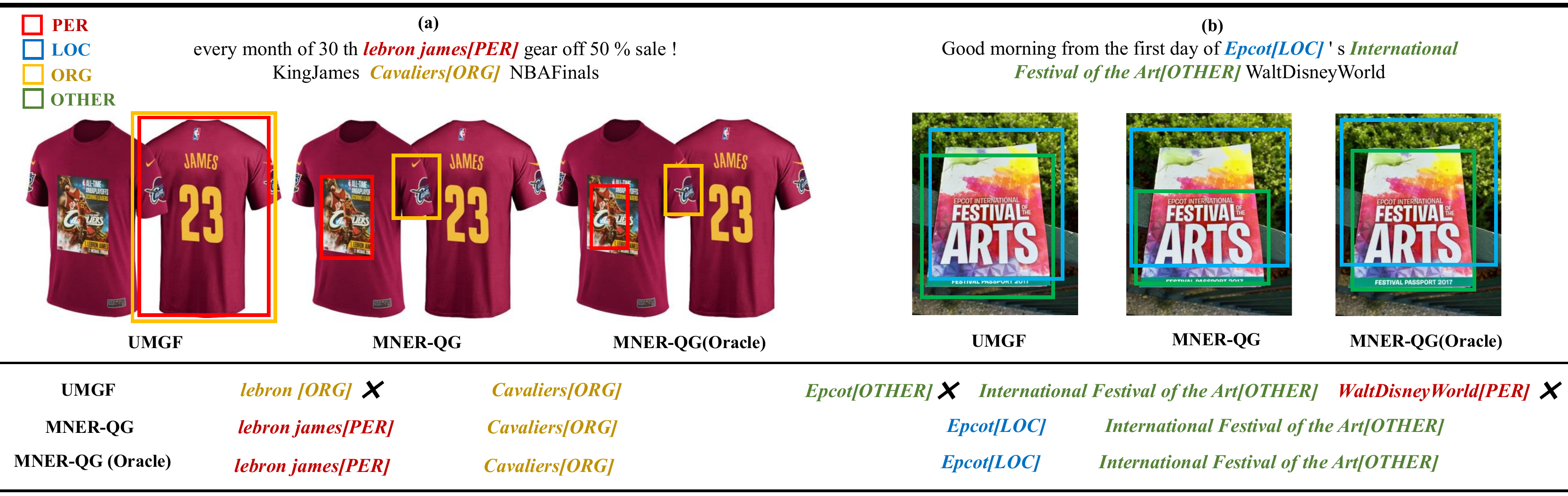}
  \caption{Example comparison among MNER-QG, MNER-QG (Oracle), and UMGF.}
  \label{fig:case_study}
\end{figure*}

\subsection{Discussions}
\label{Further Analysis}

\subsubsection{Effectiveness of the End-to-End Manner.}

\begingroup
\setlength{\tabcolsep}{3pt}
\renewcommand{\arraystretch}{0.9}
\begin{table*}[ht]
\small
\centering
\begin{tabular}{l|cccccc|cccccc}
\hline
\multicolumn{1}{c|}{\multirow{3}{*}{Methods}} & \multicolumn{6}{c|}{Twitter2015} & \multicolumn{6}{c}{Twitter2017} \\ \cline{2-13} 
\multicolumn{1}{c|}{} & \multicolumn{3}{c|}{MNER} & \multicolumn{3}{c|}{QG} & \multicolumn{3}{c|}{MNER} & \multicolumn{3}{c}{QG} \\
\multicolumn{1}{c|}{} & \textit{Pre.} & \textit{Rec.} & \multicolumn{1}{c|}{\textit{F1}} & Accu@0.5 & Accu@0.75 & Miou & \textit{Pre.} & \textit{Rec.} & \multicolumn{1}{c|}{\textit{F1}} & Accu@0.5 & Accu@0.75 & Miou \\ \hline

\multicolumn{1}{c|}{MNER-QG-Text} & \multicolumn{1}{l}{76.35} & \multicolumn{1}{l}{69.46} & \multicolumn{1}{l|}{72.74} & - & - & - & \multicolumn{1}{l}{87.12} & \multicolumn{1}{l}{84.03} & \multicolumn{1}{l|}{85.55} & - & - & - \\
\multicolumn{1}{c|}{MNER-VG} & 77.03 & 71.08 &  \multicolumn{1}{c|}{73.94} & - & - & - & 87.91 & 84.22 & \multicolumn{1}{c|}{86.03} & - & - & - \\ \hline
\multicolumn{1}{c|}{FA-VG} & - & - & \multicolumn{1}{c|}{-} & 50.83 & 32.69 & 45.49 & - & - & \multicolumn{1}{c|}{-} & 56.03 & 38.92 & 51.14 \\ \hline
\multicolumn{1}{c|}{MNER-QG (Ours)} & 77.43 & 72.15 & \multicolumn{1}{c|}{74.70} & \begin{tabular}[c]{@{}c@{}}53.93\\ (M:54.86)\end{tabular} & \begin{tabular}[c]{@{}c@{}}40.22\\ (M:41.13)\end{tabular} & \begin{tabular}[c]{@{}c@{}}49.50\\ (M:50.41)\end{tabular} & 88.26 & 85.65 & \multicolumn{1}{c|}{86.94} & \begin{tabular}[c]{@{}c@{}}57.50\\ (M:58.49)\end{tabular} & \begin{tabular}[c]{@{}c@{}}43.03\\ (M:43.67)\end{tabular} & \begin{tabular}[c]{@{}c@{}}54.09\\ (M:55.3)\end{tabular} \\ \hline
\end{tabular}
\caption{Performance comparison on Joint-training or single-training models on Test set (M denotes Max).}
\label{sub:Necessity of End-to-End Manner}
\end{table*}
Table \ref{sub:Necessity of End-to-End Manner} shows the results of our joint-training approach with other single-training approaches on different tasks\footnote{We provide two results in the QG task, one is the QG results when MNER reaches the optimum, and another is the optimal results in the QG task.}.
MNER-VG is a two-stage MNER model, which uses the VG model trained via transfer learning to acquire visual region in the first stage and integrates it into the second stage to enhance token representation. FA-VG is a one-stage VG model, and we retrain the model using Twitter2015/2017 datasets. As can be seen, compared with models MNER-QG-Text and FA-VG trained on a single data source (e.g., text or image) in different tasks, our joint-training model significantly improves the performance of each task, e.g., $F1$ score and $Accu@0.5$ are improved by 1.96\% and 3.1\% (max:4.03\%), respectively in Twitter2015.
Compared with the two-stage model MNER-VG, our end-to-end model still has obvious advantages, e.g., $F1$ scores are increased by 0.76\% and 0.91\% in Twitter2015/2017, respectively. The above results indicate that the different tasks in our model are complementary with each other under an end-to-end manner and enable the model to yield better performance.

\subsubsection{Accuracy of QG.}
\begingroup
\setlength{\tabcolsep}{2.5pt}
\begin{table}[ht]
\small
\centering
\begin{tabular}{c|cc|cc|c}
\hline
\multirow{3}{*}{Methods} & \multicolumn{2}{c|}{Twitter2015} & \multicolumn{2}{c|}{Twitter2017} & \multirow{2}{*}{Flickr30K} \\
 & (W.S) & (M.A) & (W.S) & (M.A) &  \\ \cline{2-6} 
 & A@0.5 & A@0.5 & A@0.5 & A@0.5 & A@0.5 \\ \hline
FA-VG & 50.83 & 63.94 & 56.03 & 71.02 & 68.69 \\
MNER-QG (Ours) & \begin{tabular}[c]{@{}c@{}}54.86\end{tabular} & 67.41 & \begin{tabular}[c]{@{}c@{}}58.49 \end{tabular} & 73.53 & - \\ \hline
\end{tabular}
\caption{Results on different bounding box labels on Test set (W.S and M.A denote weak supervisions and manual annotations, respectively. A@0.5 is Accu@0.5. The result of FA-VG on Flickr30K derives from \citet{yang2019fast}.)}
\label{sub:the accuracy of QG}
\end{table}
To check the quality of the labels contributed by us for the QG, we present the results of the different models on two types of labels. In addition, we provide the result on a high-quality Flickr30K Entities dataset for comparison. The dataset links 31,783 images in Flickr30K with 427K referred entities. Table \ref{sub:the accuracy of QG} shows the results. 
In either MNER-QG or FA-VG, there is a relatively obvious advantage on manual annotations compared with weak supervisions. But the acquisition of weak supervisions is easier and time-friendly. Regardless of the annotation method, our joint-training MNER-QG significantly improves the performance compared with single-training FA-VG on QG task. Compared with the results of FA-VG on Flickr30K Entities, there are competitive results on Twitter2015/2017. In particular, $Accu@0.5$ in Twitter2017 with manual annotations is 2.33\% higher than the result in Flickr30K Entities\footnote{There are great deviations in the number of images and the distribution of data in Twitter2015/2017 and Flickr30K Entities, and the comparison of the three datasets is shown in the Appendix.}. The results indicate two types of labels on Twitter2015/2017 for QG are reliable and leave ample scope for future research.

\subsubsection{Effect of Query Transformations.}
We explore different ways of query transformations and use entity type ORG as examples.
1) \textit{Keyword:} An entity type keyword. e.g.,``Organization''. 2) \textit{Rule-based Template Filling:} Phrases generated by a simple template. e.g.,``Please find Organization''. 3) \textit{Keyword's Wikipedia:} The definition of the entity type keyword from Wikipedia. e.g.,``An organization is an entity, such as an institution or an association, that has a collective goal and is linked to an external environment.'' 4) \textit{Keyword+Annotation:} The concatenation of a keyword and its annotation. e.g.,``Organization: Include club, company, government party, school...''. Results are shown in Figure \ref{fig:query tranformation}.
Queries designed by methods 1 and 2 contain deficient information, which results in friendly QG result but limits the language understanding of MRC. For method 3, definitions from Wikipedia are relatively general, leading to inferior results on both tasks. 
Compared with other methods, the framework with queries designed by method 4 achieves the highest $F1$ score and $\mathrm{Accu@0.5}$ in both tasks. We conjecture that method 4 accords with the requirements for query construction.
\begin{figure}[ht]
\begin{center}
\vspace{-3mm}
\includegraphics[width=1\linewidth,height = 0.4\linewidth]{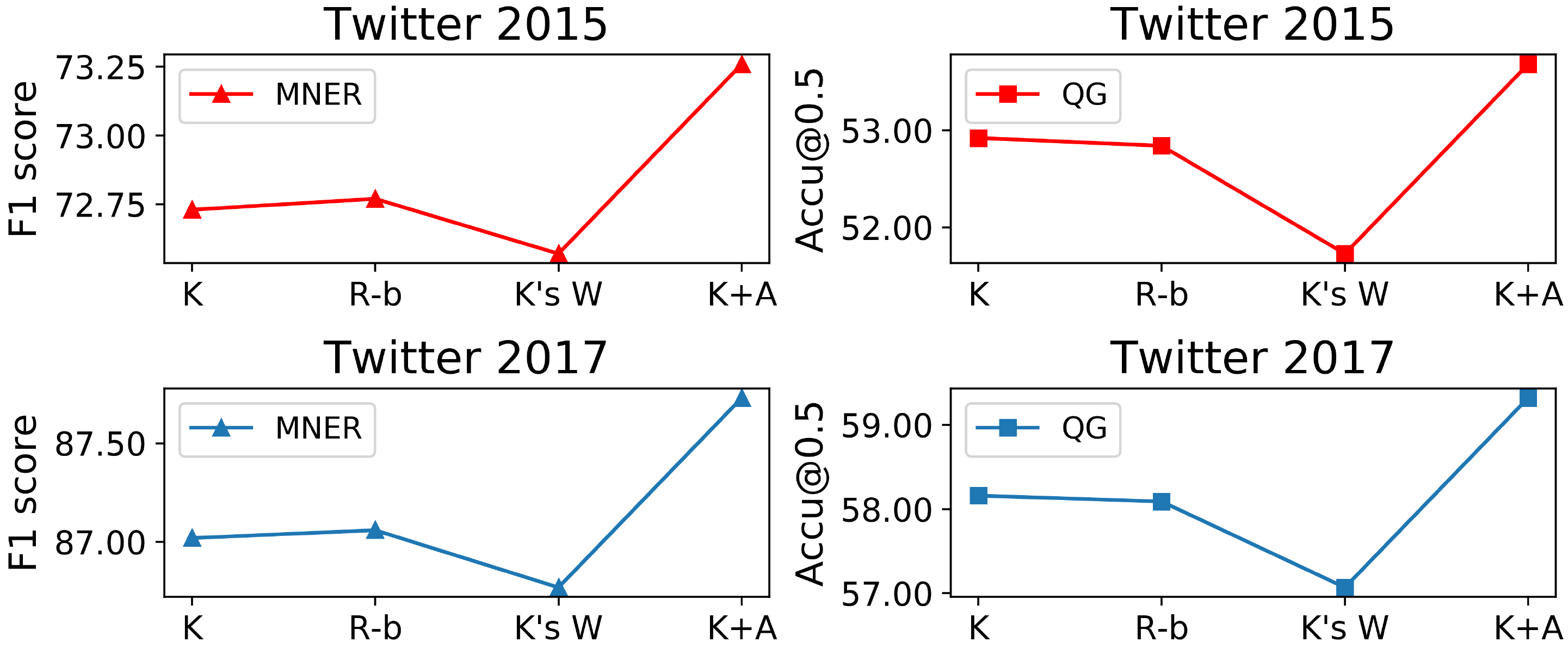}
\end{center}
\vspace{-2mm}
  \caption{Results with different query transformations in MNER and QG on Validation set (K, R-b, K's W, and K+A correspond to methods 1-4 of query transformations).}
\label{fig:query tranformation}
\end{figure}

\section{Conclusion and Future Work}
In this work, we propose MNER-QG, an end-to-end MRC framework for MNER with QG. Our model provides prior knowledge of entity types and visual regions with the guidance of queries, then enhances representations of both texts and images after the interactions of multi-scale cross-modality and existence-aware uni-modality, at last, simultaneously extracts entity span and grounds the queries onto visual regions of the image. To perform the query grounding task, we contribute weak supervisions and manual labels. 
Experimental results on two widely-used datasets show that the joint-training model MNER-QG competes strongly with other baselines in different tasks. 
MNER-QG leaves ample scope for further research. For future work, we will explore more multimodal topics.
\section{Acknowledgement}

This work is supported by the National Key R\&D Program of China under Grant No. 2020AAA0106600.

\bibliography{aaai23}

\bigskip
\pagebreak
\appendix
\section{Details of Label Attention Network}
Label attention network has been exploited to better facilitate the task of text classification \cite{socher2013zero,zhang2016zero}. Inspired by previous works \cite{cui2019hierarchically,qin2021co,modularized}, we perform the label attention network to obtain the start/end label-enhanced textual representations and label-enhanced existence representation in Existence-aware Uni-modality Interaction Module. Due to space limitation, we omit the implementation details in our paper. Here, we introduce the computational details.

\subsection{Label Representation.} Given a set of candidates' labels $\mathrm{Y}_{start}=\left \{ y_{s_{0}},...,y_{\left|start\right|-1}\right\}$, $\mathrm{Y}_{end}=\left \{ y_{e_{0}},...,y_{\left|end\right|-1}\right\} $, $\mathrm{Y}_{exist}=\left \{ y_{i_{0}},...,y_{\left|exist\right|-1}\right\} $ for start position prediction, end position prediction and existence detection, here we take the start position prediction as an example to illustrate the calculation process. Each label $y_{s_{i}}$ of start position is represented by using an embedding vector:
\begin{equation}
\mathbf{S}_{i}=\mathbf{e}^{l}\left ( y_{s_{i}} \right )  
\end{equation}
where $\mathbf{e}^{l}$ denotes a label embedding lookup table. Label embeddings are randomly initialized and tuned during model training. The label representation of start position is $\mathbf{S}\in\mathbb{R}^{\left | start \right |\times d} $, $d$ denotes the dimension of hidden state. $\left | start \right |$ represents the number of start positions. 


\subsection{Label-enhanced Textual Representation.} For the label attention layer, the attention mechanism produces an attention matrix containing the distribution of potential labels for each token. We first receive the contextual representation $\mathbf{H}_{u}$ that interacts with the image information and initial global existence representation $\mathbf{H}_{g}$. And then, we define $\mathbf{Q} =\mathbf{H}_{u}$, $\mathbf{K}=\mathbf{V}=\mathbf{S}$, and the start representation of entity span with start position label is calculated by: 
\begin{equation}
\mathbf{H}^{l}_{s}=\mathrm{softmax}\left (\frac{\mathbf{Q}\mathbf{K}^{\top}}{\sqrt{d} } \right )\mathbf{V} 
\end{equation}

We use multi-head attention to capture multiple of possible label distributions in parallel.

\begin{equation}
\mathbf{H}^{l}_{s} = \mathrm{concat}\left (\mathrm{head} ,...,\mathrm{head} _{h}\right )
\end{equation}
\begin{equation}
\mathrm{head}_{i} = \mathrm{attention}\left (\mathbf{Q}\mathbf{W}^{Q}_{i}, \mathbf{K}\mathbf{W}^{K}_{i}, \mathbf{V}\mathbf{W}^{V}_{i}\right )      
\end{equation}
where $\mathbf{W}^{Q}_{i},\mathbf{W}^{K}_{i},\mathbf{W}^{V}_{i}\in\mathbb{R}^{d\times\frac{d}{h}}$ are parameters to be learned during the training, $h$ is the number of heads.

Finally, we obtain the start label-enhanced textual representation as follows:
\begin{equation}
\mathbf{H}_{s} = \mathbf{W}^{l}_{s} \left[\mathbf{H}^{l}_{s};\mathbf{H}_{u} \right]
\end{equation}
where $\mathbf{W}^{l}_{s}\in\mathbb{R}^{d\times 2d}$ and $\mathbf{H}_{s}\in\mathbb{R}^{c\times d}$.

The calculation process of end representation of entity span and existence representation are similar to the above procedure. The end representation of entity span is $\mathbf{H}_{e} \in \mathbb{R}^{c\times d}$ and existence representation is $\mathbf{\widehat{H}}_{g} \in \mathbb{R}^{1\times d}$.

\section{Details of Datasets}


\subsection{Statistics of Datasets.} Both Twitter2015 and Twitter2017 contain four entity types: Person (PER), Organization (ORG), Location (LOC) and Others (OTHER) for sentence-image pairs. The statistics are shown in Tabel \ref{tab:Twitter distri}.

\begingroup
\setlength{\tabcolsep}{5pt}
\begin{table}[htb]
\small
\centering
\begin{tabular}{c|ccc|ccc}
\hline
\multirow{2}{*}{\textbf{Type}} & \multicolumn{3}{c|}{Twitter2015} & \multicolumn{3}{c}{Twitter2017} \\
 &
  \multicolumn{1}{l}{\textbf{Train}} &
  \multicolumn{1}{l}{\textbf{Dev}} &
  \multicolumn{1}{l|}{\textbf{Test}} &
  \multicolumn{1}{l}{\textbf{Train}} &
  \multicolumn{1}{l}{\textbf{Dev}} &
  \multicolumn{1}{l}{\textbf{Test}} \\ \hline
PER                   & 2,217     & 522       & 1,816     & 2,943     & 626       & 621      \\
LOC                   & 2,091     & 522       & 1,697     & 731       & 173       & 178      \\
ORG                   & 928       & 247       & 839       & 1,674     & 375       & 395      \\
OTHER                  & 940       & 225       & 726       & 701       & 150       & 157      \\ \hline
Total                 & 6,176     & 1,546     & 5,078     & 6,049     & 1,324     & 1,351    \\ \hline
\textbf{\# Tweets}    & 4,000     & 1,000     & 3,257     & 3,373     & 723       & 723      \\ \hline
\end{tabular}
\caption{The statistics summary of two MNER datasets.}
\label{tab:Twitter distri}
\end{table}

\subsection{Details of Weak Supervisions.} 

We train a visual grounding model via transfer learning to offer weak supervisions. 
To fulfill MNER-QG adaptation of the pre-trained FA-VG model, we construct a corpus consisting of three sets of samples:
\begin{enumerate}
\item Samples from the Flickr30K Entities dataset with phrases highly-related to pre-defined PER, LOC, ORG, and OTHER queries.
\item Samples from (1) with phrases replaced by MNER-QG queries.
\item Samples from the training set of Twitter2015/2017 datasets with manually-labeled regions of PER, LOC, and ORG types.
\end{enumerate}

Since only a small part of phrases in the Flickr30K Entities dataset are related to MNER-QG entity types, we first filter the original data to get those highly-relevant samples, and replace phrases in them with MNER-QG queries. Specifically, all phrases in the Flickr30K Entities dataset and four MNER-QG queries are represented by BERT embeddings respectively. Then we calculate cosine similarities between embeddings of phrases and each query, and only keep samples with scores larger than a threshold, e.g., 0.7. To obtain samples in the second set, we make a copy of the first set and conduct query replacement. 
\begin{figure}[ht]
\centering
  \includegraphics[scale=0.56]{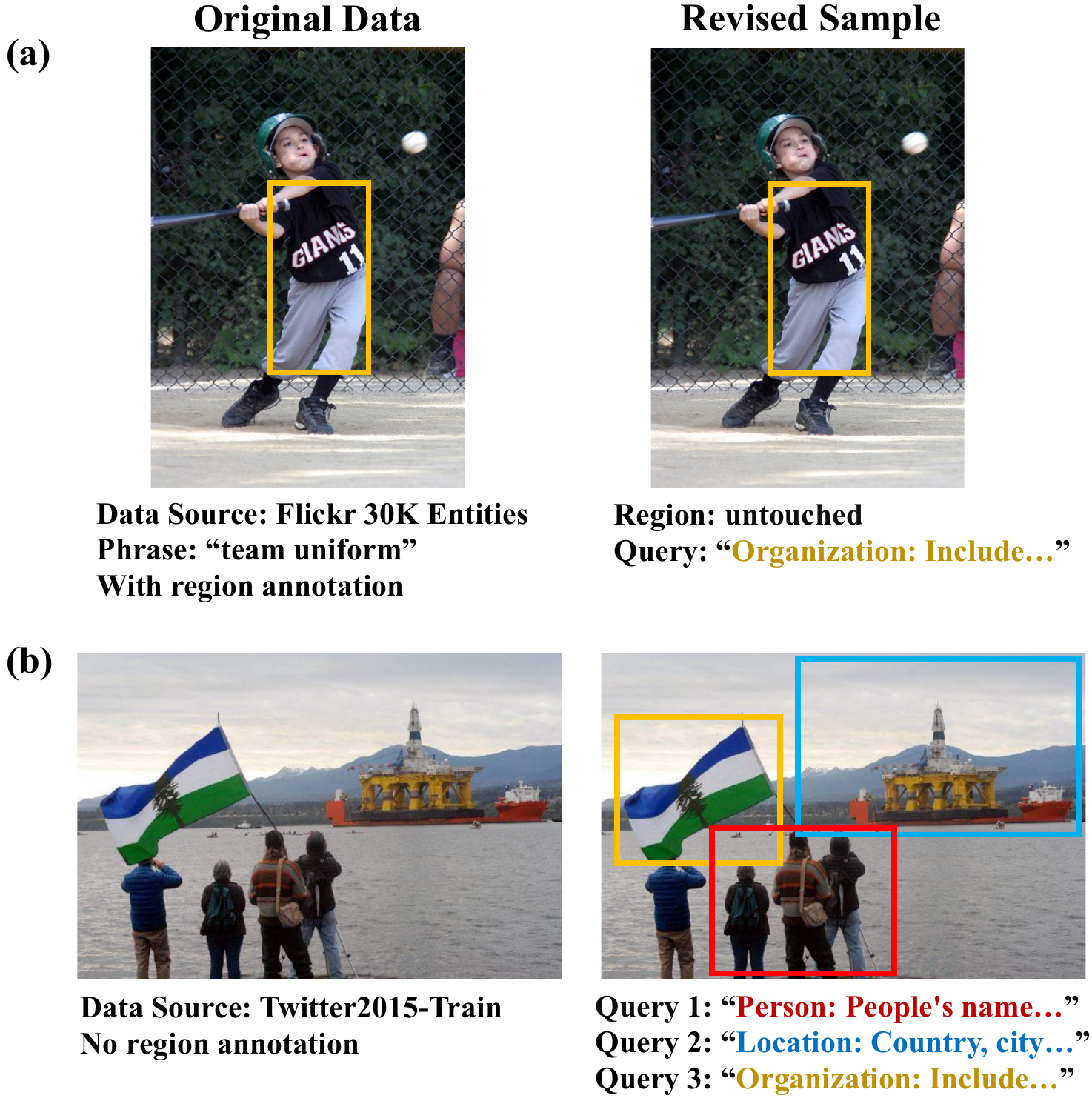}
  \caption{Illustration of Corpus Construction. (a) Replacing original phrases in the Flickr30K Entities dataset with MRC queries. (b) Labelling existing regions related to PER, LOC, and ORG in the Twitter2015/2017-Training dataset.}
  \label{fig:data_revision}
\end{figure}
As shown in Figure \ref{fig:data_revision}(a), the original query ``team uniform'' is replaced by ``Organization: Include club, company...'' which is defined as the MNER-QG query for ORG in the table \textit{Examples for transforming entity types to queries} of this paper. To take advantage of some in-domain data, we randomly sample 1000+ images from the \textbf{training set} of Twitter2015/2017 datasets, and manually annotate regions related to PER, LOC, ORG, and OTHER. Take Figure \ref{fig:data_revision}(b) as an example. The image is labeled with three pairs of regions and queries: red box with ``Person: People's name...'', blue box with “Location: Country, city...'' and yellow box with “Organization: Include...''. The statistics of the constructed corpus are summarized in Table \ref{tab:statistics of the constructed corpu}. We split the corpus into training/validation/test set with the ratio of 9:0.5:0.5. During training, all sets of samples are shuffled so that the model can be finetuned to not only maintain the ability of accurate visual grounding, but also adapt to new task and domain. 

\begingroup
\setlength{\tabcolsep}{1.8pt}
\renewcommand{\arraystretch}{1}
\begin{table}[htp!]
\small
\centering
\begin{tabular}{r|l}
\hline
Total data volume                                 & 26,311                                 \\
F.30K data (unmodified)                       & 12,504                                 \\
F.30K data + modified query data              & 12,504                                 \\
Tw.15/17 data + query + b.-box & 1,303                                   \\
LOC query data                                    & 2,983 (F.30K) + 700 (Tw.15/17) \\
ORG query data                                    & 4,191 (F.30K) + 350 (Tw.15/17) \\
PER query data                                    & 4,362 (F.30K) + 253 (Tw.15/17) \\
OTHER query data                                   & 968 (F.30K)                        \\ \hline
\end{tabular}
\caption{Statistics of our constructed VG corpus (F.30k and Tw.15/17 denote Flickr30k and Twitter2015/2017, respectively and b.-box denotes bounding box).}
\label{tab:statistics of the constructed corpu}
\end{table}

FA-VG makes predictions based on the fused representations of image, text, and spatial\footnote{The spatial feature captures the coordinates of the top-left corner, center, and bottom-right corner of the grid at ($i$, $j$).} features. After being finetuned on the constructed corpus, FA-VG achieves a grounding accuracy (IoU$>$0.5) of 79.96\% on the test set\footnote{IOU (Intersection over Union) is a term used to describe the extent of overlap of two boxes.}, which significantly outperforms the pre-trained FA-VG model without MNER adaptation (64.72\%). The results show that the highly flexible visual grounding model with transfer learning can reach a better localization of regions related to entity types. Finally, we obtain a set of bounding boxes labels from the output layer of FA-VG.

\subsection{Details of Manual Annotations.}
Three crowd-sourced workers are hired to annotate the bounding boxes in the image of Twitter2015 and Twitter2017. We ask crowd-sourced workers to match sentences and images in Tweets when they perform annotations. If a sentence does not contain entities of a particular type, the labeled bounding boxes should be covered a whole image. We assign 4.8k Tweets to each crowd-sourced worker from a total of 13K+ Tweets in two datasets and collect 1.3k cross-annotated Tweets. The agreement between annotators is measured using a percentage of 50\% IoU overlapping for image regions. To ensure the quality of ground-truth, we follow the previous works \cite{bowman2015large,chen2019tabfact,jia2022convrec} to employ the Fleiss Kappa \cite{fleiss1971measuring} as an indicator, where Fleiss $\mathcal{K}=\frac{\bar{p}_{c}-\bar{p}_{e}}{1-\bar{p}_{e}}$ is calculated from the observed agreement $\bar{p}_{c}$ and the agreement by chance $\bar{p}_{e}$. We obtain a Fleiss $\mathcal{K}=0.85$, which indicates strong inter-annotator agreement.

\subsection{Comparisons of Twitter2015/2017 and Flickr 30K Entities.} Here, we present the comparisons of Twitter2015/2017 and Flickr 30K Entities in terms of images' and queries' distributions. The details are shown in Table \ref{tab:comparison of TWitter and Flickr}. We refer to the statistics of images in the Flickr 30K Entities dataset from \citet{plummer2015flickr30k}. The distribution of queries in the Flickr 30K Entities is obtained by counting the number of queries in the dataset\footnote{We acquire the Flickr 30K Entities dataset on the website \url{https://bryanplummer.com/Flickr30kEntities/}.}. The queries in Twitter2015/2017 are designed by us and not included in the original dataset. From Table \ref{tab:comparison of TWitter and Flickr}, we can see that the number of both images and queries in the Flickr30K Entities is more than the number in Twitter2015/2017.

\begingroup
\setlength{\tabcolsep}{0.8pt}
\renewcommand{\arraystretch}{1.2}
\begin{table}[ht]
\small
\centering
\begin{tabular}{c|ccc|ccc|ccc}
\hline
\multirow{2}{*}{Nums} & \multicolumn{3}{c|}{Twitter2015} & \multicolumn{3}{c|}{Twitter2017} & \multicolumn{3}{c}{Flickr 30K} \\
 & Train & Dev & Test & Train & Dev & Test & Train & Dev & Test \\ \hline
Image & \multicolumn{1}{c|}{4,000} & \multicolumn{1}{c|}{1,000} & 3,257 & \multicolumn{1}{c|}{3,373} & \multicolumn{1}{c|}{723} & 723 & \multicolumn{1}{c|}{29,783} & \multicolumn{1}{c|}{1,000} & 1,000 \\ \hline
Query & \multicolumn{1}{c|}{16,000} & \multicolumn{1}{c|}{4,000} & 13,028 & \multicolumn{1}{c|}{13,492} & \multicolumn{1}{c|}{2,892} & 2,892 & \multicolumn{1}{c|}{69,888} & \multicolumn{1}{c|}{5,066} & 5,016 \\ \hline
\end{tabular}
\caption{Comparisons of Twitter2015/2017 and Flickr 30K Entities.}
\label{tab:comparison of TWitter and Flickr}
\end{table}

\section{Details of Experiment Settings}

\subsection{Details of Baseline Models.}
\begin{figure*}[ht]
\centering
  \includegraphics[scale=0.52]{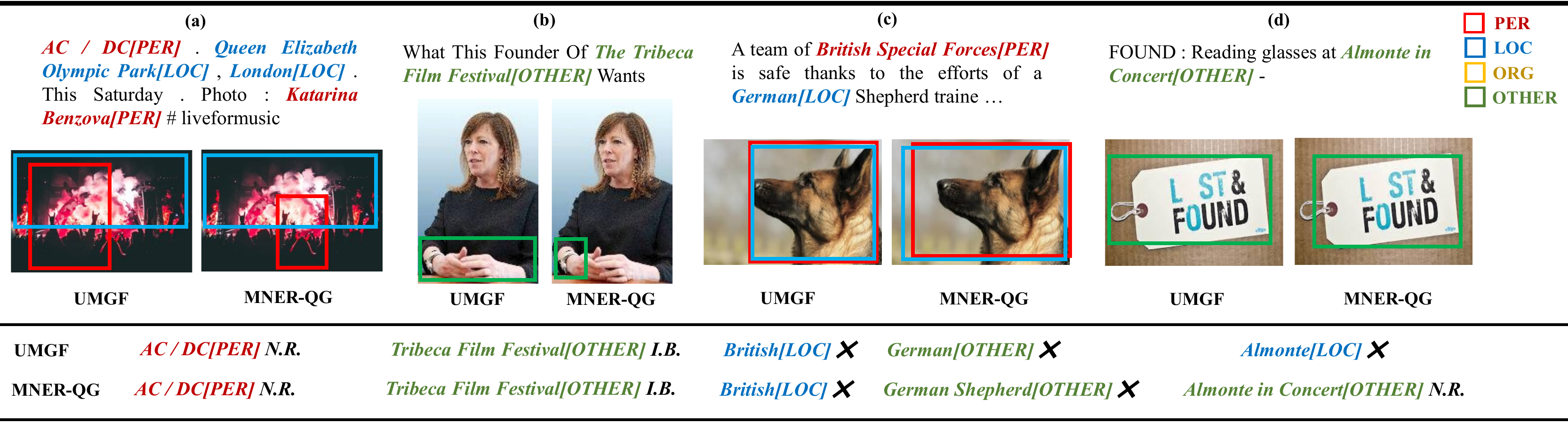}
  \caption{Error Cases by our model MNER-QG and UMGF (N.R. and I.B. denote Not Recognized and Inaccurate Boundaries, respectively).}
  \label{fig:error_case}
\vspace{-2mm}
\end{figure*}
We compare two groups of baselines with our approach. The first group consists of some text-based NER models that formalize NER as a sequence labeling task:
(1) \textbf{BiLSTM-CRF}, which is the vanilla NER model with a bidirectional LSTM layer and a CRF layer. (2) \textbf{CNN-BiLSTM-CRF}, which is an improvement of BiLSTM-CRF. The embedding of each word is substituted with the concatenation of word-level embedding and character embedding. (3) \textbf{HBiLSTM-CRF}, which is a variant of CNN-BiLSTM-CRF by replacing the CNN layer with the LSTM layer to acquire the character embedding. (4) \textbf{BERT}, which is a competitive model for NER with a multi-layer bidirectional Transformer encoder followed by a softmax layer. (5) \textbf{BERT-CRF}, which is a variant of BERT by replacing the softmax layer with a CRF decoder. (6) \textbf{T-NER}, which is a NER model designed specifically for tweets. It exploits widely-used features, including the dictionary, contextual and orthographic features.

Besides, we compare several competitive multimodal NER models: (1) \textbf{GVATT-HBiLSTM-CRF}, which uses HBiLSTM-CRF to encode text and proposes a visual attention mechanism to acquire text-aware image representation. (2) \textbf{GVATT-BERT-CRF}, which is a variant of the GVATT-HBiLSTM-CRF by replacing the text encoder HBiLSTM with BERT. (3) \textbf{AdaCAN-CNN-BiLSTM-CRF}, which is the MNER model based on CNN-BiLSTM-CRF, designing an adaptive co-attention mechanism to integrate image and text. (4) \textbf{AdaCAN-BERT-CRF}, which is a variant of the AdaCAN-CNN-BiLSTM-CRF by replacing the text encoder CNN-BiLSTM with BERT. (5) \textbf{UMT-BERT-CRF}, which proposes a multimodal interaction module to acquire expressive text-visual representation by incorporating the auxiliary entity span detection into multimodal Transformer. (6) \textbf{MT-BERT-CRF}, which is a variant of the UMT-BERT-CRF, pruning the auxiliary module. (7) \textbf{ATTR-MMKG-MNER}, which integrates both image attributes and image knowledge into the MNER model. (8) \textbf{UMGF}, which proposes graph fusion approach based on graph model to obtain text-visual representation. (9) \textbf{MAF}, which proposes a matching and alignment framework for MNER to alleviate the impact of mismatched text-image pairs on encoding.

\subsubsection{Setting of Individual Parameter.}
In the joint-training process, we find there is a very large gap between QG loss and MNER loss. We specially design a balance factor $\omega_{f}$ to dynamically scale the losses of two tasks to the same order of magnitude. The equation is as follows:

\begin{equation}
\omega_{f}=\frac{1}{10^{ \left | \left \lfloor\log_{10}{a}\right \rfloor- \left \lfloor\log_{10}{b}\right \rfloor   \right | }}
\end{equation}
where $\left \lfloor  \right \rfloor $ denotes the $\mathrm{floor()}$ function. $\left \lfloor x \right \rfloor=\mathrm{max}\left \{ n\in \mathbb{Z} |n\le x\right \} $. $a$ and $b$ represent losses in different tasks. For example, in this paper, $a$ is one loss in the Entity Span Prediction sub-task (e.g., $\mathcal{L}_{start}$), $b$ is the loss $\mathcal{L}_{QG}$ in the Query Grounding task.

\section{Further Discussions}

\subsection{Error Analyses.} 
Here, we present four representative error cases to fulfill further analyses, meanwhile, we append the results of UMGF as a comparison. The entities and entities' types in the sentence are highlighted by italicized bold fonts in different colors.

In Figure \ref{fig:error_case}(a), the sentence contains four entities ``\textit{AC / DC}'', ``\textit{Katarina Benzova}'', and ``\textit{Queen Elizabeth Olympic Park}'', ``\textit{London}'' with PER, and LOC types respectively. We can observe that neither UMGF nor MNER-QG recognizes ``\textit{AC / DC}'' although they detect visual regions of the PER type in the image. We conjecture that this is because the image provides only obscure information and ``\textit{AC / DC}'' does not have a clear semantic for PER type. Figure \ref{fig:error_case}(b) shows a difficult case, the sentence contains one entity ``\textit{The Tribeca Film Festival}'' with OTHER type. There is only one woman wearing a watch and no regions for OTHER type in the image. Both UMGF and MNER-QG recognize entity ``\textit{Tribeca Film Festival}'' with OTHER type but the entity cannot exactly match with ``\textit{The Tribeca Film Festival}''. We guess that models can locate the entity span associated with the OTHER type but can not detect the exact entity boundary. Figure \ref{fig:error_case}(c) is a more challenging case, the sentence contains two entities ``\textit{British Special Forces}'' and ``\textit{German}'' with PER and LOC types respectively. Here, the entity ``\textit{British Special Forces}'' should be recognized as ORG type, not PER. Meanwhile, another entity ``\textit{British}'' with LOC type is embedded in ``\textit{British Special Forces}''. Hence, both UMGF and MNER-QG recognize the entity ``\textit{British}'' with LOC type but ``\textit{British}'' is not the target in this case. The image contains the region of ``\textit{German Shepherd}'', our model MNER-QG grasps this visual clue and recognizes the entity ``\textit{German Shepherd}'' with OTHER type but ``\textit{German Shepherd}'' is still not the target in this case. UMGF recognizes the entity ``\textit{German}'' but mis-recognizes the type of the entity as OTHER. The entity that should be recognized in Figure \ref{fig:error_case}(d) is ``\textit{Almonte in Concert}'' and its corresponding entity type is OTHER. Similar to the aforementioned cases, the image is still not informative. Hence, our model ignores the entity ``\textit{Almonte in Concert}'', and UMGF mis-recognizes the entity ``\textit{Almonte}'' with LOC type.

As we can see from the above error cases, most images cannot provide effective visual clues for the text message. The result indicates that informative visual clues in the image are important for multimodal named entity recognition. At the same time, we also find that our model has a high error rate on OTHER type. This result reminds us to pay more attention to the OTHER type. The aforementioned four challenging cases reveal that high-quality datasets are a hinge in MNER and we still have a long way to go for this task.
 
\end{document}